# Coarse and fine-grained automatic cropping deep convolutional neural network

Jingfei Chang

School of Computer Science and Information Engineering, Hefei University of Technology, Hefei 230009, China

## Abstract

The existing convolutional neural network pruning algorithms can be divided into two categories: coarse-grained clipping and fine-grained clipping. This paper proposes a coarse and fine-grained automatic pruning algorithm, which can achieve more efficient and accurate compression acceleration for convolutional neural networks. First, cluster the intermediate feature maps of the convolutional neural network to obtain the network structure after coarse-grained clipping, and then use the particle swarm optimization algorithm to iteratively search and optimize the structure. Finally, the optimal network tailoring substructure is obtained.

## Keywords：

## 1 Introduction

Deep neural networks [1][2][3] have achieved great success in scientific research and engineering with the continuous expansion of data sets and the substantial increase in hardware computing power. As one of the important branches, convolutional neural network [4] has outstanding performance in image feature extraction by virtue of its parameter sharing and translation invariance characteristics. It has been widely used in image classification [5], target detection [6], style Significant breakthroughs have been made in the fields of transformation [7] and semantic segmentation [8]. As image tasks become more complex, the depth and width of the network are gradually increasing, and the scale of the network has also become extremely large while improving performance. As a result, some existing deep convolutional neural networks can only be trained on GPUs, TPUs, or cloud processors, and cannot be deployed on mobile terminals and wearable devices with limited computing and storage capabilities and high real-time requirements. It greatly limits the development and application of convolutional neural networks.

In order to solve this problem, a large number of researchers have compressed and accelerated the deep convolutional neural network, aiming at the redundancy of the parameters and structure of the existing network, reducing the number of parameters and the number of floating point operations (FLOPs). The current mainstream methods mainly include designing dedicated hardware architecture [9] [10], optimizing convolution calculation methods [11] [12] and designing network compression algorithms. Among them, network compression algorithms are mainly divided into four types: network pruning, low-rank factorization [13], quantification [14] [15] [16] and knowledge distillation [17]. The low-rank factorization operation can achieve network sparseness and directly compress and accelerate the network. However, additional calculations are introduced in the implementation process, which is not conducive to the reduction of floating-point operations. Quantization can speed up the calculation but cannot reduce the number of parameters, and when the network is more complex, the inference accuracy is relatively low. Knowledge distillation can make the deeper model narrower, but the similarity requirements for the two model tasks are higher, and it may get worse performance in actual use. The pruning algorithm of the convolutional neural network starts from the redundancy of the network, and uses the importance evaluation index of the parameters to cut the unimportant parameters and their connections. The pruning algorithm has received widespread attention because it is easy to implement and can effectively compress and accelerate the original convolutional neural network. Existing convolutional neural network pruning algorithms can be roughly divided into two categories: the first category is coarse-grained clipping, such as based on the convolution kernel and the magnitude of the channel [18] [19][20] or based on similarity [21] The implementation process of this type of method is relatively simple, but the stability is poor, and the final compact network experiment has a large loss of accuracy; the second type is fine-grained pruning, for example, [22] considers error propagation for pruning By minimizing the final response layer, the neurons are clipped layer by layer to obtain a compact network. [23] implements network pruning through reinforcement learning, etc. The final compact network obtained by this type of method has less accuracy loss, but the cutting process is time-consuming Labor intensive. [21] proved the effectiveness of using the cosine similarity between feature maps as a parameter redundancy evaluation index. [24] demonstrated the guiding role of neural structure search in neural network structure optimization. Inspired by the above two works, this paper proposes a coarse and fine-grained automatic pruning algorithm.

    The coarse and fine-grained automatic pruning algorithm in this paper first performs inference tests on the pre-trained baseline convolutional neural network on the image classification data set, statistics the feature maps generated during the test, and calculates the cosine distance between each two feature maps , And then perform cluster analysis on the number of channels in each layer of the network according to this distance, and use the clustering result as the number of channels after coarse-grained cropping in each layer to form a convolutional neural network structure after coarse-grained cropping. Then initialize the network structure on the basis of the initially formed compact network structure, and finally use the particle swarm optimization algorithm to automatically search and optimize these network structures to obtain the final compact neural network to realize the compression and acceleration of the original convolutional neural network . Coarse-grained clustering and cropping of the original network first, and then fine-grained cropping through automatic iterative search optimization has the following two advantages: First, the coarse-grained clustering and cropping itself is efficient, and it can provide smaller size for fine-grained cropping. Initialize the network structure range of the network, accelerate the efficiency of fine-grained search and tailor, and then make the entire network tailoring process more efficient; secondly, coarse-grained tailoring can provide a better initialization network structure for the optimization algorithm during fine-grained tailoring, thereby reducing The accuracy of the final compact network is lost. This article has carried out concrete experimental verification on three data sets of CIFAR-10, CIFAR-100 and ILSVRC-2012.

The coarse and fine-grained automatic pruning algorithm in this article can directly perform channel-level pruning on the existing convolutional neural network structure without introducing sparsity, without the assistance of additional sparse matrix operations and acceleration libraries, and the entire pruning process is It is realized by controlling the hyperparameters, which greatly reduces the intervention of manual operation, and can realize the automatic compression and acceleration of the network.

The main contributions of this article are as follows:

1. This paper proposes a coarse-grained pruning algorithm for convolutional neural networks, which clusters all channels in each layer of the original network, and uses the sum of the number of clusters and the number of noise channels as the current layer The number of channels after preliminary pruning, after clustering all layers, the coarse-grained pruned structure of the original convolutional neural network is obtained.

2. This paper proposes a fine-grained pruning algorithm for convolutional neural networks. The network structure obtained by coarse-grained pruning is initialized as a structure population, and then the particle swarm optimization algorithm is used to iteratively search and optimize the coarse-grained pruning. The obtained compact network structure is optimized to obtain the optimal pruning network.

3. This article combines coarse-grained and fine-grained pruning algorithms to propose a coarse and fine-grained automatic pruning algorithm for convolutional neural networks. Only one hyperparameter is controlled to achieve different amplitude compression and acceleration of the original network, which is easy to implement and can make the final result. The compact network has a high image classification accuracy.

4. This article conducted a lot of comparative experiments on the CIFAR-10, CIFAR-100 and ILSVRC-2012 data sets, and proved the effectiveness and accuracy of the coarse and fine-grained automatic pruning algorithm in the compression and acceleration of convolutional neural networks. The ablation analysis experiment verified that the adjustment of hyperparameters has a stable control on the network clipping ratio.

The latter part of this article is organized as follows: The second part introduces related work in this field; the third part specifically describes the coarse and fine-grained pruning method proposed in this article and its implementation details; the fourth part shows a large number of comparative verification experiments and ablation analysis; The full text is summarized.

## 2 Related work

Deep convolutional neural networks have made major breakthroughs in various fields of images, but their huge parameters and floating point operations limit the network's operation on mobile terminals and wearable devices. At present, there has been a lot of work devoted to network compression and acceleration. Among them, the convolutional neural network pruning algorithm has attracted widespread attention due to its simple implementation and obvious effects, and many excellent algorithms have emerged. The existing network pruning algorithms can be roughly divided into two categories: coarse-grained network pruning and fine-grained network pruning.

**Coarse-grained network pruning**

Coarse-grained clipping refers to from the perspective of the importance of parameters and connections, according to the proposed importance evaluation index, the unimportant parameters and connections in the network are directly deleted, and then the obtained compact network is fine-tuned or retrained to restore the experimental accuracy. The core of this type of algorithm is to design a reasonable and effective parameter importance evaluation index. [4] The scaling factor of the network batch normalization layer is used as the evaluation index of the channel importance, and the unimportant channels are deleted to realize the channel-level network pruning. [18] used the size of the weight as an index to evaluate the importance of connections, and deleted all connections in the trained network whose weights were lower than the threshold. [20] First, calculate and sort the L1 norm of the convolution kernel, and cut all the convolution kernels with smaller values and their corresponding feature maps according to the preset clipping ratio. [25] regards the L2 norm as the evaluation index of the importance of the convolution kernel, and continuously updates the convolution kernel that was cropped last time during the training phase. [26] The fully connected layer is used as a linear classifier to extract the feature representation of the middle layer of the network, and the layer with less improved feature representation will be clipped according to a predefined threshold. [27] According to the Euclidean distance between the convolution kernel in each layer and the geometric mean of all convolution kernels in that layer, the redundancy of the convolution kernel is judged, and the redundancy in the network layer is reduced according to the set clipping rate. The convolution kernel is cropped. [28] introduced connection sensitivity to evaluate the importance of structural connections, and tailored the network during the parameter initialization stage before training. [29] Considering the clipped layer and the neuron connections before and after it, the neural network is compressed by minimizing the F norm caused by multi-layer clipping. [30] It is proposed to use the rank of the feature map to determine how much information it contains. The low-rank feature map contains less information and can be cut with confidence. The high-rank feature map contains more information, even if a fixed part is not updated. The final model performance will not have a big impact. [21] proposed a new convolution kernel pruning method, which uses the diversity and similarity of feature maps to cut redundant and similar feature maps and their corresponding convolution kernels in the network.

**Fine-grained network pruning**

Fine-grained tailoring refers to considering the impact of tailoring strategy on network accuracy loss during the network tailoring process, tailoring the network layer by layer, and continuously adjusting the tailoring strategy during the tailoring process. The core of this type of algorithm is to design a reasonable and effective network tailoring adjustment strategy. [22] Considering the score propagation of the importance of neurons, using feature sorting technology to measure the importance of neurons in the penultimate layer in front of the classification layer, and tailoring the network by minimizing the reconstruction error of this layer. [31] introduced the variational technique to estimate the channel saliency in the training process, looked for a suitable probability distribution to model it, and finally cut the redundant channel according to the model parameters. [32] proposed a feature enhancement and suppression method. The significantly important channels are pre-enhanced at runtime and the unimportant channels are skipped. This method retains the complete network structure. [33] For the pre-trained network, each layer provides multiple tailoring standards, allowing the network to learn by itself, and finally determine the tailoring standards for each layer. Then retrain the cropped network to restore accuracy. [34] Add discrete gates after the convolutional layer, and control the deletion of channels in the convolutional layer by training discrete gates. [35] proposed differentiable Markov channel clipping, which cuts the original network by searching for the optimal substructure, and optimizes it by cross-entropy loss and gradient descent constrained by floating-point operands. [36] Combining channel pruning and model fine-tuning into a single end-to-end training process, an effective channel selection layer is proposed, which automatically finds less important convolution kernels by means of joint training, and this selection layer will The activation response is taken as input, and a binary index code is generated to delete the convolution kernel. [37] proposed a network channel pruning scheme based on sparse learning and genetic algorithm.

## 3 Method

This paper proposes a coarse and fine-grained automatic pruning algorithm, which aims to achieve more efficient and accurate channel-level compression and acceleration of the existing convolutional neural network, thereby reducing the number of network parameters and floating point operations, so that it can be used in calculations Deploy on embedded devices with limited capacity and power consumption. In this part, the overall framework of the coarse and fine-grained automatic pruning algorithm is first proposed, and then the two main components of the algorithm are specifically introduced: coarse-grained clustering and pruning and fine-grained structure search optimization.

## 3.1 Overall framework of coarse and fine-grained automatic pruning algorithm

Given a pre-trained original convolutional neural network structure and a set of pictures randomly sampled from the image classification data set, the sampled images are first inferred on the original network, and the feature maps obtained in each layer are integrated and clustered. Achieve coarse-grained tailoring of the network structure. Then the obtained network structure is initialized as a structure population, and each network is searched and optimized for a limited number of iterations using the training data set. The redundant channels and their connected convolution kernels are deleted, and finally the most comprehensive cropped Excellent convolutional neural network structure.

## 3.2 Coarse-grained clustering

Given a layer of convolutional neural network, for a data set $D = \{(x_1, y_1), (x_2, y_2), \ldots, (x_m, y_m)\}$, where $x_i$ is the input data and $y_i$ is the label, the convolution operation is parameterized as $Conv = \mathcal{F}(D, W; C)$, where $C = (c_1, c_2, \ldots, c_L)$ is the original network structure, $c_i$ is the number of channels in each layer, $W \in \mathbb{R}^{n \times c \times k \times k}$ is the convolution kernel, n is the number of output channels and c is the number of input channels and $k \times k$ is the size of the convolution kernel. The feature map generated by the convolution operation is $M \in \mathbb{R}^{S \times c \times W \times H}$, where $S$ is the number of samples sampled by the convolution operation and $W \times H$ is the feature map size.

The task of this part is to use the clustering algorithm to delete the redundant feature maps and their corresponding convolution kernels in each layer to realize the coarse-grained clipping of the convolutional neural network. For a single input image, the feature map generated by each convolutional layer is $M_i \in \mathbb{R}^{c_i \times W_i \times H_i}$, and then this $c_i$ channel is clustered, and the dimension of each channel is $W_i \times H_i$. In this paper, a density-based clustering algorithm (DBSCAN) using noise is used [38][39]. The algorithm defines the classified clusters as the largest collection of connected points, which can divide the regions with sufficiently high density into clusters, and can find clusters of arbitrary shapes in the spatial database containing noise. Specify the number of final clusters.

First, set the neighborhood parameters of clustering $(\varepsilon, MinPts)$. The similarity between two feature maps $M_i$ and $M_j$ in the same layer is calculated by cosine distance, which is defined as follows:

$$dis(M_{ij}) = \left| \cos\left[ \left( \frac{1}{S} \sum_{s=1}^{S} M_i^s \right), \left( \frac{1}{S} \sum_{s=1}^{S} M_j^s \right) \right] \right|_\circ \tag{1}$$

If the $\varepsilon$-neighborhood of $M_i$ contains at least $MinPts$ feature maps, that is $|\mathcal{N}_\varepsilon(M_i)| \geq MinPts$, then classify $M_i$ as a core feature map, and all core feature maps form a set. Randomly select a feature map in the set. All feature maps in the $\varepsilon$-neighborhood are classified as a cluster. If some feature maps in the $\varepsilon$-neighborhood are also core feature maps, then these core feature maps-feature maps in the $\varepsilon$-neighborhood Also belong to the cluster. Neither the core feature map nor any core feature map-feature maps in the neighborhood are classified as noise feature maps. According to this clustering method, the number of channels in each layer of the network is clustered by controlling the neighborhood parameters, and the sum of the number of clusters and the number of noises obtained by clustering is used as the final number of clustering channels, and then the convolution is obtained. The coarse-grained tailoring structure of the neural network $C' = (c_1', c_2', \ldots, c_L')$.

## 3.3 Fine-grained structure search optimization

The previous section performed a preliminary coarse-grained tailoring of the original convolutional neural network structure $C$. In this section, we further carry out precise and detailed structure search and optimization on the tailored network structure $C'$. In order to achieve this goal, we use particle swarm optimization algorithm [40][41] to optimize the structure $C'$ and find the compact network structure $C^*$ with the largest test accuracy, which satisfies:

$$C^* = \arg\max_{C'} Acc\left( D_{Test}\left( D_{Train}, W'; C' \right) \right), \tag{2}$$

Among them, $D_{Train}$ is the training data set and $D_{Test}$ is the test data set. Fine-grained structure search optimization can be divided into the following two steps.

**Network structure initialization**
First, initialize the particle swarm according to the network structure $C'$ obtained by coarse-grained tailoring to obtain a set of $N$ initial network structures $\{C_i^0\}$, among which $i \in [1, N]$. In order to make the initial network structure similar, and the number of channels in each layer has a certain change, to achieve the purpose of random initialization. This paper proposes an initialization method for the network structure after coarse-grained tailoring. The specific implementation is as follows:

$$c_{il}^0 = c_l' + i \times rand\{-1, 0, 1\}, \tag{3}$$

Among them, the structure of the i-th initial network is $C_i^0 = \left(c_{i1}^0, c_{i2}^0, \ldots, c_{iL}^0\right)$, $rand\{-1, 0, 1\}$ means randomly sampling a number among -1, 0, and 1.

Then, initialize the channel update speed collection $\{V_i^0\}$ in the network structure search process, where $V_i^0 = \left(v_{i1}^0, v_{i2}^0, \ldots, v_{iL}^0\right)$, set the maximum update speed to $v_{max}$, $v_i^0 \in [-v_{max}, v_{max}]$.

The fitness of $C_i^0$ is calculated according to formula 4:

$$fitness(C_i) = \max Acc\left(D_{Test}\left(D_{Train}, W_i; C_i\right)\right), \tag{4}$$

In the entire process of fine-grained structure search and optimization, there are many calculations of fitness, and network training is time-consuming. When calculating the fitness of each network structure, this article only trains a few cycles, and then performs it on the test set. Test, use the test result as the adaptability of the network structure. This can not only reflect the adaptability of the network structure but also shorten the structure search time. Initialize each network structure as the local optimal network structure, and use the structure with the largest fitness as the initialization of the global optimal network structure $pbest$.

**Iterative search optimization**

After the initialization is completed, iterative search is started to find the optimal convolutional neural network structure. First update the search speed of the channel, the specific update method is as follows:

$$v_i^t = w \times v_i^{t-1} + \alpha_1 \times rand \times \left(pbest_i - c_i^{t-1}\right) + \alpha_2 \times rand \times \left(gbest_i - c_i^{t-1}\right), \tag{5}$$

Among them, $v_i^t$ represents the channel change speed of the i-th network structure in the t-th iteration, $\alpha_1$ and $\alpha_2$ are the learning factor. This article uses the recommended value $\alpha_1 = \alpha_2 = 2$ in [41] because it can make the weights in "society" and "cognition" an average of 1, $rand$ is a random number between 0 and 1, and $w$ is the inertia factor. Since the dynamic $w$ can obtain better search and optimization results than the fixed value, the linear incremental weight strategy is adopted in this article, and the specific update method of $w$ is as follows:

$$w^t = \left(w_{ini} - w_{snd}\right)(T - t)/T + w_{snd}, \tag{6}$$

Where $w_{ini}$ is the initial inertia factor, $w_{snd}$ is the inertia factor when iterating to the maximum update algebra, $T$ is the maximum network search iteration number, and $t$ is the current iteration number. Then update the candidate network substructure $C_i^t$ as follows:

$$c_i^t = c_i^{t-1} + r \times v_i^t, \qquad (7)$$

Among them, $r$ is the learning rate, in this article $r = 2$.

Then, calculate the fitness of each candidate substructure according to formula 4. If it is greater than the fitness of the initial local optimal structure, update the candidate substructure to the new local optimal structure $pbest_i = C_i^t$, if its fitness is greater than the global optimal structure, then update the global optimal structure $gbest = pbest_i$. Finally, after $T$ sub-network structure iterative search, the global optimal structure obtained is the network after the final tailoring. Algorithm 1 shows the pseudo-code for cropping the pre-trained network in this article. Given a pre-trained convolutional neural network, crop it according to Algorithm 1, to obtain the compressed compact network $C^*$, and then perform retraining and recovery experiments accuracy.

---

**Algorithm 1** PSO pruning algorithm

---

**Input:** Search cycles:$T$, Numbers of pruned structure:$N$, Max velocity:$max\_vel$
**Output:** Optimal compact structure: $(C')^*$
 1: **for** $i = 1$ to $N$ **do**
 2:     initialize the pruned structure set according to the clustering result: $\{C_i^0\}$;
 3:     initialize the velocity set: $\{V_i^0\}$;
 4:     calculate the fitness of $C_i$ via Eq0: $fitness\,(C_i^0)$ ;
 5:     $pbest_i = C_i^0$;
 6: **end for**
 7: $gbest = \max\{pbest_i\}$;
 8: **for** $t = 1$ to $T$ **do**
 9:     **for** $i = 1$ to $N$ **do**
10:         update the search velocity of channels via Eq0: $V_i^t$;
11:         update the candidate structure via Eq0: $C_i^t$;
12:         calculate the fitness of candidate structure via Eq0: $fitness\,(C_i^t)$;
13:         **if** $fitness\,(C_i^t) > fitness\,(pbest_i)$ **then**
14:             $pbest_i = C_i^0$;
15:             $fitness\,(pbest_i) = fitness\,(C_i^t)$;
16:         **end if**
17:         **if** $fitness\,(pbest_i) > fitness\,(gbest)$ **then**
18:             $gbest = pbest_i$;
19:             $fitness\,(gbest) = fitness\,(pbest_i)$;
20:         **end if**
21:     **end for**
22: **end for**
$(C')^* = gbest$.

---

For the cut network, most of the existing methods inherit the convolution kernel weights and biases of the trained original network, and fine-tune to restore the performance of the original network as much as possible. However, when the network clipping range is large and the parameter loss is large, the performance recovery after fine-tuning is not obvious, and the actual performance of the compact network after pruning cannot be fully demonstrated. [42] revealed a surprising feature in structured network clipping: using inherited weights to fine-tune the clipped model will not be better than training the clipped model directly from scratch, and the final result is proved through experiments The cropped structure (relative to the weights of the original network trained) may be more important.

In this paper, the results of experiments using VGG-16 on the CIFAR-10 data set further verify the correctness of the conclusions in [35]. In order to fully demonstrate the performance of the tailored network, follow-up experiments adopt a retraining method from scratch for compact networks. The training method from scratch that keeps the number of floating-point operations (FLOPs) consistent is adopted. Specifically, the multiple of FLOPs compression is multiplied by the number of training cycles of the original network as the number of retraining cycles of the cropped network, and the obtained network accuracy is compared with the original Compare the uncut network and draw a conclusion.

# 4 Experiments

In order to verify the effectiveness of the method proposed in this paper, we conducted experiments on VGGNet, ResNet, and GoogleNet on the CIFAR-10 and CIFAR-100 datasets, and cut and compressed ResNet on the ILSVRC-2012[43] dataset. The experimental platform chooses Facebook's open source pytorch framework. In order to obtain better benchmark accuracy and be able to make a fair comparison with some existing network pruning methods, the pre-training and hyperparameter setting of the network use the method proposed in [44], and the initial learning rate in the training phase is set to 0.1, Reduce by 10 times at the half cycle and three quarter cycle respectively, and use momentum-driven stochastic gradient descent (SGD) for back propagation, set the momentum coefficient to 0.9, add the L2 regular term to the final loss function, and set the coefficient Is 0.0001.

## 4.1 Datasets

The CIFAR-10 data set contains 60000 RGB color images, a total of 10 categories, each category contains 5000 training images, 1000 test images. On the CIFAR-10 dataset, we experimented with four networks: VGG-16, ResNet-56/110 and GoogleNet.

The CIFAR-100 data set contains 60000 RGB color images, a total of 100 categories, each category contains 600 images, including 500 training images and 100 test images. On CIFAR-100, we experimented with three networks: VGG-19 and ResNet-56/110.

The ILSVRC-2012 data set is a large-scale image classification data set with a total of 1,000 categories, including approximately 1.2 million training images, 50,000 verification images and 150,000 test images. On ILSVRC-2012, we experimented with ResNet-18/34/50 three networks.

## 4.2 VGG-16 on CIFAR-10

| Dataset | Model | Acc/% | Acc.drop/% | Parameters | Parameters.drop/% | FLOPs | FLOPs.drop/% |
|---|---|---|---|---|---|---|---|
| CIFAR-10 | VGG-16 | 93.60 | - | 14.73M | - | 314.59M | - |

|  | Base | | | | | |
|---|---|---|---|---|---|---|
|  | 0.01 0, 5 | 94.03 | -0.43 | 5.02M | 65.72% | 152.64M | 51.48% |
|  | 0.02 0, 5 | 93.66 | -0.06 | 2.76M | 81.28% | 93.52M | 70.25% |
|  | 0.03 5, 5 | 93.45 | 0.15 | 1.23M | 91.66% | 83.54M | 73.44% |

# 5 Conclusion

This paper proposes a coarse and fine-grained pruning algorithm for convolutional neural networks. For a given convolutional neural network, firstly perform coarse-grained clustering and cropping, and then use particle swarm optimization algorithm to perform fine-grained optimization. Get the final compact network. The experiment proves that our method can make the pruning process of convolutional neural network more efficient and accurate.